\documentclass[11pt]{article}

\usepackage{hyperref}
\usepackage{url}

\usepackage{amsmath,amsbsy,amsfonts,amssymb,amsthm,dsfont}
\usepackage{cite,xspace}
\usepackage{algorithm,algorithmic}
\usepackage{graphicx}
\usepackage{rotating}

\usepackage{color}
\usepackage{url}
\usepackage{float}
\usepackage{fullpage}
\usepackage{bm}

\definecolor{darkblue}{rgb}{0.0,0.0,0.5}

\newcommand{\system}[1]{\textsf{\footnotesize #1}}

\newcommand{\errortop}{{\ensuremath\varepsilon}}
\newcommand{\errorbot}{{\ensuremath\epsilon}}
\newcommand{\regrettop}{{\ensuremath r}}
\newcommand{\regretbot}{{\ensuremath \rho}}
\newcommand{\regret}{\operatorname{reg}}
\newcommand{\argmax}{\operatornamewithlimits{arg\,max}}
\newcommand{\I}{{\mathbf{I}}}
\newcommand{\E}{{\mathbf{E}}}
\newcommand{\hide}[1]{}

\title{Learning Reductions that Really Work}

\author{Alina Beygelzimer\\ Yahoo Labs \\ beygel@yahoo-inc.com
\and
Hal Daum\'{e} III \\ University of Maryland, College Park\\ hal@umiacs.umd.edu 
\and 
John Langford \\ Microsoft Research\\ jcl@microsoft.com 
\and 
Paul Mineiro \\ Microsoft CISL\\ pmineiro@microsoft.com}

\begin{document}

\maketitle
\begin{abstract}
We provide a summary of the mathematical and computational techniques
that have enabled learning reductions to effectively address a wide
class of problems, and show that this approach to
solving machine learning problems can be broadly useful.
\end{abstract}

\section{Introduction}

In a reduction, a complex problem is decomposed into simpler
subproblems so that a solution to the simpler subproblems gives
a solution to the complex problem. When this is a simple process, it is
conventionally called ``programming'', while ``reduction'' is reserved
for more difficult applications of this technique. Computational
complexity theory, for example, relies on reductions in an essential
fashion to define computational complexity classes, most canonically
NP-hard problems. In machine learning, reductions are often used
constructively to build good solutions to hard problems.

\begin{figure}[h]
\centering
\includegraphics[width=3.5in]{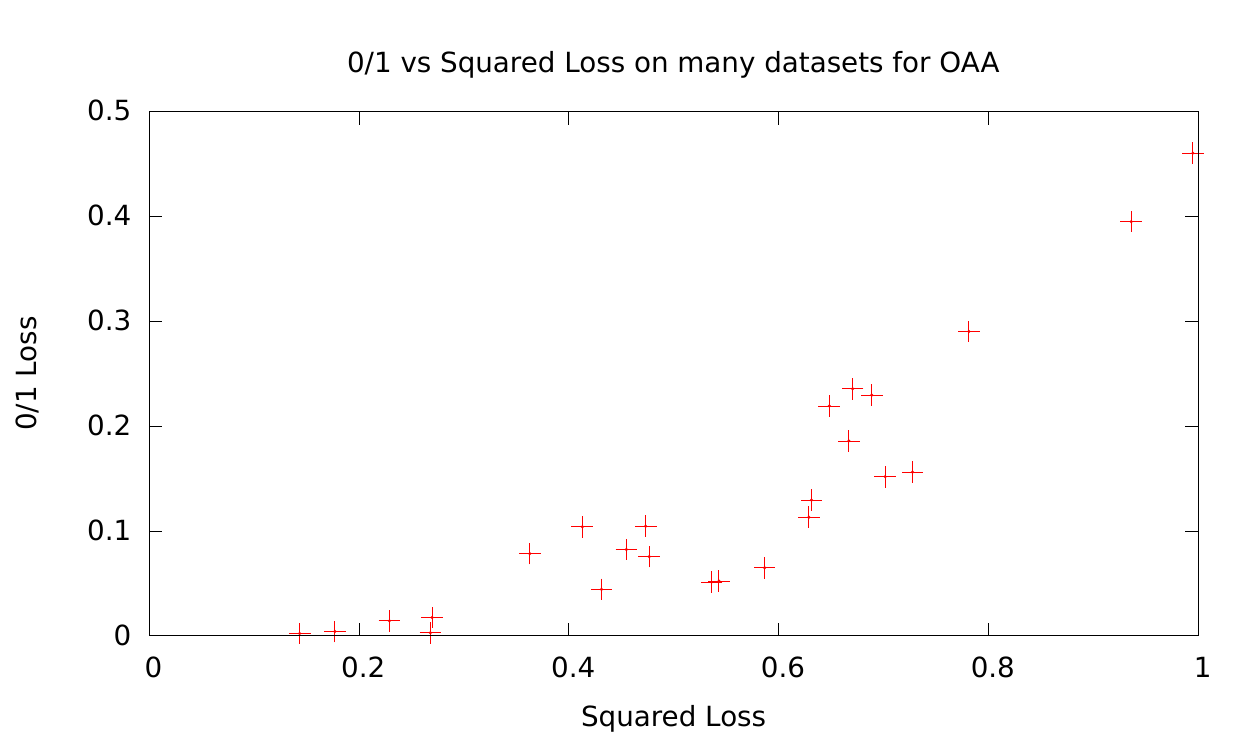}
\caption{Multiclass loss rate compared to average base predictor loss
  rate for the one-against-all reduction applied to 2-class
  classification datasets. (Normally, one-against-all
is applied to problems with more
  than 2 classes.  However, we expect the relative scaling of base
  predictor loss and multiclass loss to vary with the number of
  classes, so it was desirable to fix the number of classes for this
  plot.)}
\label{fig:oaa}
\end{figure}

The canonical example here is the one-against-all reduction, which
solves $k$-way multiclass classification via reduction to $k$
base predictions:  The $i$th predictor is trained to predict whether a
label is class $i$ or not. Figure~\ref{fig:oaa} shows how this
reduction works experimentally, comparing the multiclass loss to the
average loss of base predictors.  There are no base predictors with
small loss inducing a large multiclass loss---as ruled out by theory.

For this paper, a learning reduction takes as input complex examples,
transforms them into simpler examples, invokes an appropriate learning
algorithm on the simpler examples, then transforms predictions on
these simpler examples to a prediction on the complex
examples. Learning reductions vary in the definition of ``complex
examples'', ``simpler examples'', and ``transforms''.  

Are learning reductions an effective approach for solving complex
machine learning problems?  The answer is not obvious, because there
is a nontrivial representational concern: maybe the process of
reduction creates ``hard'' problems that simply cannot be solved well?
A simple example is given by 3-class classification with a single
feature and a linear predictor.  If class $i$ has feature value $i$,
then the $2$ versus $1,3$ classifier necessarily has a large error
rate, causing the one-against-all reduction to perform poorly.  
The all-pairs reduction, which learns a classifier for each
pair of labels, does not suffer from this problem in this case.

Although this concern is significant, it is of unclear strength as
there are many computationally convenient choices made in machine
learning, such as conjugate priors, proxy losses, and sigmoid link
functions.  Perhaps the representations created by natural learning
reductions work well on natural problems?  Or perhaps there is a
theory of representation respecting learning reductions?

We have investigated this approach to machine learning for about a
decade now, and provide a summary of results here, addressing
several important desiderata:
\begin{enumerate}
\item A well-founded theory for analysis. A well-founded theory makes
  the approach teachable, and provides a form of assurance that good
  empirical results should be expected, and carry over to new
  problems.
\item Good statistical performance. In addition, the theory should
  provide some effective guidance about which learning algorithms are
  better than other learning algorithms. 
\item Good computational performance. Computational limitations in
  machine learning are often active, particularly when there are large
  amounts of data. This is critical for learning reductions, because
  the large data regime is where sound algorithmics begin to
  outperform clever representation and problem understanding.
\item Good programmability. Programmability is a nontraditional concern
  for machine learning which can matter significantly in practice.
\item A unique ability. Learning reductions must either significantly
  exceed the performance of existing systems on existing problems or
  provide a means to address an entirely new class of problems for the
  effort of mastering the approach to be justified.
\end{enumerate}
Here we show that all the above criteria have now been met.

\subsection{Strawman one-against-all}

A common approach to implementing 
one-against-all for $k$-way multiclass classification 
is to create a script that processes the dataset $k$ times, creating $k$
intermediate binary datasets, then executes a binary learning algorithm
$k$ times, creating $k$ different model files. For test time evaluation,
another script then invokes a testing system $k$ times for each example
in a batch. The multiclass prediction is the label with a positive prediction, 
with ties broken arbitrarily. 

A careful study of learning reductions reveals that every aspect of
this strawman approach can be improved. 

\subsection{Organization}

Section~\ref{sec:theory} discusses the kinds of reduction theory that 
have been developed and found most useful.

Section~\ref{sec:interface} discusses the programming interface
we have developed for learning reductions.  Although programmability
is a nonstandard concern in machine learning applications, we have
found it of critical importance.  Creating a usable
interface which is not computationally constraining is critical to
success.

Section~\ref{sec:unique} discusses several problems for which the only
known solution is derived via a reduction mechanism, providing
evidence that the reduction approach is useful for research.

Section~\ref{sec:l2s} shows experimental results for a particularly
complex ``deep'' reduction for structured prediction, including comparisons
with many other approaches.

Together, these sections show that learning reductions are a useful
approach to research in machine learning.

\section{Reductions theory}
\label{sec:theory}
There are several natural learning reduction theories on a spectrum
from easy to powerful, which we discuss in turn.

\subsection{Error reductions}
In an error reduction, a small error rate on the created problems
implies a small error rate on the original problem. When multiple base
problems are created, we measure the average error rate over the
base problems. Since it is easy to resample examples or indicate
via an importance weight that one example is more important than the
other, nonuniform averages are allowed.

For example, in the strawman one-against-all reduction, 
an average binary classification
error rate of $\epsilon$ implies a multiclass error rate of at most
$(k-1)\epsilon$ (see \cite{allwein,guruswami-sahai}).  
A careful examination of the analysis shows how to improve 
the error-transformation properties of the reduction: 
The first observation is that it helps to break ties randomly instead 
of arbitrarily.  The second observation is that, in the absence of other errors, a false negative 
implies only a $1/k$ probability of making the right multiclass prediction,
while for a false positive this probability is $1/2$.
Thus modifying the reduction to make the binary classifier more prone 
to output a positive, which can be done via an appropriate use of 
importance weighting, improves the error transform from $(k-1)\epsilon$ 
to roughly $\frac{k}{2}\epsilon$. 
As predicted by theory, 
both of these elements yield an improvement in practice~\cite{WOAA}.

Another example of an error reduction for multiclass classification is
based on error-correcting output codes~\cite{ecoc,guruswami-sahai}.

A valid criticism of error reductions is that the guarantees they provide 
become vacuous if the base problems they create are inherently noisy.
For example, when no base binary classifier can achieve
an error rate better than $\frac{2}{k}$, the one-against-all guarantee
above is vacuous.

\subsection{Regret reductions}
Regret analysis addresses this criticism by analyzing the transformation of 
\emph{excess} loss, or \emph{regret}.  
Here regret of a predictor is the difference between its loss and 
the minimum achievable loss on the same problem.
In contrast to many other forms of learning theory, the minimum is
over \emph{all} predictors.

A reduction that translates 
any optimal (i.e., no-regret) solution to the base problems 
into an optimal solution to the top-level problem is called \emph{consistent}.
Consistency is a basic requirement for a good reduction.
Unfortunately, error reductions are generally inconsistent.
To see that one-against-all is inconsistent, 
consider three classes with true conditional probabilities 
$\frac{1}{2}-2\delta$, 
$\frac{1}{4}+\delta$, and $\frac{1}{4}+\delta$.  
The optimal base binary prediction is always
0, resulting in multiclass loss of $2/3$.  The corresponding multiclass regret
is $\frac{1}{6}-2\delta$, which is positive for any $\delta < 1/12$.

\begin{figure}
\centering
\includegraphics[width=3.5in]{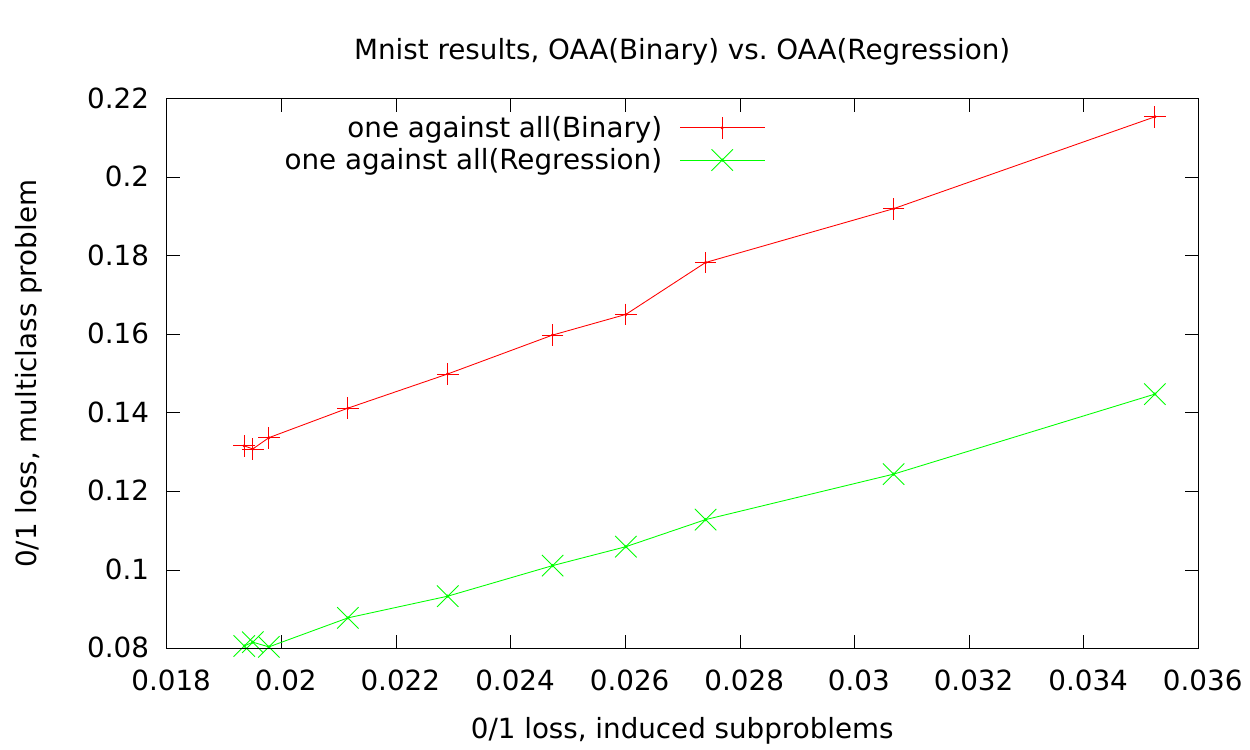}
\caption{Mnist experimental results for one-against-all reduced to
  binary classification and one-against-all reduced to squared loss
  regression while varying training set size.  The x-axis is the 0/1
  test loss of the induced subproblems, and the y-axis is the 0/1 test
  loss on the multiclass problem.  The classifiers are linear in
  pixels.  The regression approach (a regret transform) dominates the
  binary approach (an error transform).}
\label{fig:mnist}
\end{figure}

Strawman one-against-all can be easily made consistent by reducing to 
squared-loss regression instead of binary classification.
The multiclass prediction is made by evaluating the learned regressor
on all labels and predicting with the argmax.
As shown below, this regression approach is consistent.
It also resolves ties via precision rather 
than via randomization, as seems
more likely to be effective in practice.  Figure \ref{fig:mnist}
illustrates the empirical superiority of reducing to regression rather
than binary classification for the Mnist~\cite{mnist} data set.

We will analyze the regret transform of this approach for any fixed $x$, taking expectation over $x$ at the end.  Let $f(x,a)$ be the learned regressor, predicting the conditional probability of class $a$ on $x$.
Let $p_a$ be the true conditional probability.
The squared loss regret of $f$ on the predicted 
class $a$ is 
\begin{align*}
& \E_{y\, \mid\, x}[{(f(x,a)-\I(y=a))^2-(p_a-\I(y=a))^2]}\\
= & (p_a -f(x,a))^2.
\end{align*}
Similarly, the regret of $f$ on the optimal label $a^* =\argmax_a p_a$ 
is $(p_{a^*}-f(x,a^*))^2$.  To incur multiclass regret, we must have 
$f(x,a) \geq f(x,a^*)$.  The two regrets are convex and the minima is reached when $f(x,a) = f(x,a^*) = \frac{p_a + p_{a^*}}{2}$.  The corresponding 
squared loss regret suffered by $f$ on both $a$ and $a^*$ is 
$(p_a^*-p_a)^2/2$.  Since the regressor doesn't need to incur any loss on 
other predictions, the regressor can pay $\regret(f) = (p_a^*-p_a)^2/2k$ 
in average squared loss regret
to induce multiclass regret of $p_{a^*}-p_a$ on $x$.  
Solving for multiclass regret in terms of $\regret(f)$ shows that the 
multiclass regret of this approach is bounded by 
$\sqrt {2k \regret(f)}$.
Since the adversary can play this optimal strategy the bound is tight.

Although a regret reduction is more desirable than an error reduction,
the typical square root dependence introduced when analyzing regret 
is not desirable.
Nonetheless, moving from an error reduction to a regret reduction 
is often empirically beneficial (see Figure~\ref{fig:mnist}).

\bigskip\noindent
There are many known regret reductions for such problems as
multiclass classification~\cite{picts,ramaswamy}, cost-sensitive 
classification~\cite{langford,ECT}, and ranking~\cite{ranksort,agarwal}.
There is also a rich body of work on so called \emph{surrogate regret bounds}.
It is common to use some efficiently minimizable 
surrogate loss instead of the loss one actually wishes to optimize. 
A surrogate regret bound quantifies the resulting regret in terms 
of the surrogate regret~\cite{bartlett, agarwal, reid}.
These results show that standard 
algorithms minimizing the surrogate are in fact consistent solutions
to the problem at hand. In some cases, commonly used surrogate losses 
actually turn out to be inconsistent~\cite{duchi-mackey-jordan}.

\subsection{Adaptive reductions}
Adaptive reductions create learning problems that are dependent on the
solution to other learning problems.  In general, adaptivity is
undesirable, since conditionally defined problems are more difficult
to form and solve well---they are less amenable to parallelization, and
more prone to overfitting due to propagation and compounding of errors. 

In some cases, however, the best known reduction is adaptive.
One such example is logarithmic time multiclass
prediction, discussed in section~\ref{sec:log-time}.
All known unconditional log-time approaches yield inconsistency 
in the presence on label noise~\cite{ECT}.

The average base regret is still well defined 
as long as there is a partial order over
the base problems, i.e., each base learning problem is defined
given a predictor for everything earlier in the order.

\bigskip\noindent
Boosting~\cite{BoostingBook} can be thought of as an adaptive
reduction for converting any weak learner into a strong learner.
Typical boosting statements bound the error rate of the resulting 
classifier in terms of the weighted training errors $\epsilon_t$ on 
the distributions createdy adaptively by the booster.
Ability to boost is rooted in the assumption of weak learnability---
a weak learner gets a positive edge over random guessing for any 
distribution created by the booster.
As with any reduction, there is a concern that the booster may create ``hard''
distributions, making it difficult to satisfy the assumption.
Although linear separability with a positive margin 
implies weak learnability~\cite{BoostingBook}, linear separability 
is still a strong assumption.
Despite this concern, boosting has been incredibly effective in practice.


\subsection{Optimization oracle reductions}
When the problem is efficiently gathering information, as in active
learning (discussed in section~\ref{sec:ass}) or contextual bandit
learning (discussed in section~\ref{sec:cb}), previous types of
reductions are inadequate because they lack any way to quantify
progress made by the reduction as examples are used in learning.

Suppose we have access to an oracle which when given a dataset returns
an minimum loss predictor from a class of predictors $H$ with a
limited capacity.  The form of the learning problem solved by the
oracle can be binary classification, cost-sensitive classification, or
any other reasonable primitive.  Since many supervised learning
algorithms approximate such an oracle, these reductions are
immediately implementable.

Since the capacity of $H$ is limited, tools from statistical learning
theory can be used to argue about the regret of the predictor returned
by the oracle.  Cleverly using this oracle can provide solutions to
learning problems which are exponentially more efficient than simpler
more explicit algorithms for choosing which information to
gether~\cite{minimonster,AALWoC}.

\section{Interfaces for learning reductions}
\label{sec:interface}

A good interface for learning reductions should simultaneously be performant,
generally useful, easy to program, and eliminate systemic bugs.

\subsection{The Wrong Way}
The strawman one-against-all approach illustrates interfacing failures
well.  In particular, consider an implementation where a binary learning
executable, treated as a black box, is orchestrated to do the 
one-against-all approach via shell scripting.
\begin{enumerate}
\item Scripting implies a mixed-language solution, which is relatively
  difficult to maintain or understand.
\item The approach may easily fail under recursion. For example, if
  another script invokes the one-against-all training script multiple
  times, it is easy to imagine a problem where the saved models of one
  invocation overwrites the saved models of another invocation. In a good
  programming approach, these sorts of errors should not be possible.
\item The transformation of multiclass examples into binary examples
  is separated from the transformation of binary predictions into
  multiclass predictions. This substantially raises the possibility of
  implementation bugs compared to an approach which has encoder and
  decoder implemented either side-by-side or conformally.
\item For more advanced adaptive reductions, it is common to require a
  prediction before defining the created examples. Having a prediction
  script operate separately creates a circularity (training must
  succeed for prediction to work, but prediction is needed for
  training to occur) which is extremely cumbersome to avoid in this
  fashion.
\item The training approach is computationally expensive since the
  dataset is replicated $k$ times.  Particularly when datasets are
  large, this is highly undesirable.  
\item The testing process is structurally slow, particularly when there
  is only one test example to label. The computational time is $\Omega(pk)$
  where $p$ is the number of parameters in a saved model and $k$ is
  the number of classes simply due to the overhead of loading a model. 
\item Even if all models are loaded into memory, the process of querying
  each model is inherently unfriendly to a hardware cache.
\end{enumerate}

\subsection{A Better Way}

Our approach~\cite{VW} eliminates all of the above interfacing bugs,
resulting in a system which is general, performant, and easily
programmed while eliminating bugs due to antimodular implementation.

We require a base learning algorithm which presents two
\emph{online} interfaces: 
\begin{enumerate}
\item {\verb+Predict+}(example $e$, instance $i$) returns a prediction for base problem $i$ and is guaranteed to not update the internal state.
\item {\verb+Learn+}(example $e$, instance $i$) returns the same result as {\verb Predict }, but may update the internal state.
\end{enumerate}
Note that although we require an online learning algorithm interface,
there is no constraint that online learning must occur---the manner in
which state is updated by {\verb Learn } is up to the base learning
algorithm.  The interface certainly favors online base learning algorithms,
but we have an implementation of LBFGS~\cite{LBFGS} that functions as
an effective (if typically slow) base learning algorithm.

Since reductions are composable, this interface is both a constraint
on the base learning algorithm and a constraint on the learning
reduction itself---the learning reduction must define its own
{\verb Predict} and {\verb Learn} interfaces. 

It is common for reductions to have some state which is summarized in
a reduction-specific datastructure.  \emph{Every} reduction requires a
base learner which may either be another reduction or a learning
algorithm.  Reductions also typically have some reduction-specific
state such as number of classes $k$ for the one-against-all reduction.
In a traditional object-oriented language, these arguments can be 
provided to the constructor of the reduction and encapsulated in
the reduction object.  In a purely functional language, the input 
arguments can be augmented with an additional
state variable (and the return value of {\verb Learn } augmented 
with an updated state variable).

The above interface addresses all the previously mentioned problems
except for problem (7).  Consider a dense linear model with sparse
features.  In this situation, the speed of testing is commonly limited
by the coherency of memory access due to caching effects.  To use
coherent access, we \emph{stripe} models over memory.  In particular,
for a linear layout with 4 models, the memory for model $i$ is at
address $i$, $i+4$, $i+8$,...  This layout is nearly transparent to
learning reductions---it is achieved by multiplying feature ids by the
total number of models at the top of the reduction stack, using the
instance to define offsets to feature values as an example descends
the stack, and then requiring base learning algorithms to access
example features via a foreach\_feature() function that transparently
imposes the appropriate offset for a model.

\section{Uniquely solved problems}
\label{sec:unique}
Do reductions just provide a modular alternative that performs as well 
as other, direct methods?
Or do they provide solutions to otherwise
unsolved problems?  Rephrased, are learning reductions a good first
tool for developing solutions to new problems?  

We provide evidence that the answer is `yes' by surveying an array of
learning problems which have been effectively addressed only via
reduction techniques so far.

A common theme throughout these problems is computational efficiency.
Often there are known inefficient approaches for solving intractable
problems.  Using a reduction approach, we can isolate the inefficiency
of optimization, and remove other inefficiencies, often resulting in
exponential improvements in efficiency in practice.

\subsection{Efficient Contextual Bandit Learning}
In contextual bandit learning, a learning algorithm needs to be
applied to \emph{exploration data} to learn a \emph{policy} for acting
in the world.  A \emph{policy} is functionally equivalent to a
multiclass classifier that takes as input some feature vector $x$ and
produces an action $a$.  The term ``policy'' is used here, because the
action is executed---perhaps a news story is displayed, or a medical
treatment is administered.  \emph{Exploration data} consists of quads
$(x,a,r,p)$ where $x$ is a feature vector, $a$ is an action, $r$ is a reward,
and $p$ is the probability of choosing action $a$ on $x$.

Efficient non-reduction techniques exist only for special cases of
this problem~\cite{Banditron}.  All known techniques for the general
setting~\cite{Bianca,offset,DR} use reduction approaches.

\subsection{Efficient Exploration in Contextual Bandits}
\label{sec:cb}
Effectively doing contextual bandit learning in the online setting
requires efficiently creating a good probability distribution over
actions.  There are approaches to this problem based on exponential
weights~\cite{EXP4} with a running time linear in the size of the
policy set.  Can this be done in more efficiently?

The answer turns out to be ``yes''~\cite{minimonster}.  In particular,
it is possible to reduce the problem to $O(\sqrt{T})$ instances of
cost-sensitive classification which are each trained to find
good-but-different solutions.  This is an exponential improvement in
computational complexity over the previous approach.

\subsection{Efficient Agnostic Selective Sampling}
\label{sec:ass}
A learning algorithm with the power to choose which examples to label
can be much more efficient than a learning algorithm that passively
accepts randomly labeled examples.  However, most such approaches
break down if strong assumptions about the nature of the problem are
not met.  

The canonical example is learning a threshold on the real line in the absence of any noise.
A passive learning approach requires
$O(1/\epsilon)$ samples to achieve error rate $\epsilon$, while 
selective sampling requires only $O\left(\ln
\frac{1}{\epsilon}\right)$ samples using binary search.
This exponential improvement, however, is quite brittle---a small
amount of label noise can yield an arbitrarily bad predictor.

Inefficient approaches for addressing this brittleness statistically have been
known~\cite{A2,Hanneke}.  Is it possible
to benefit from selective sampling in the agnostic setting 
efficiently?

Two algorithms have been created~\cite{AALWoC,OracularCAL} which
reduce active learning for binary classification to
importance-weighted binary classification, creating practical
algorithms.  No other efficient general approaches to agnostic
selective sampling are known.

\subsection{Logarithmic Time Classification}
\label{sec:log-time}
Most multiclass learning algorithms have time and space complexities
linear in the number of classes when testing or training.  Furthermore,
many of these approaches tend to be \emph{inconsistent} in the
presence of noise---they may predict the wrong label regardless of the
amount of data available when there is label noise.  

It is easy to note that logarithmic time classification \emph{may} be
possible since the output need only be $O(\log k)$ bits to uniquely
identify a class.  Can logarithmic time classification be done in a
consistent and robust fashion?  

Two reduction algorithms~\cite{ECT,Log_multi} provide a solution to
this.  The first shows that consistency and robustness can be achieved
with a logarithmic time approach, while the second addresses learning
of the structure directly.  

\section{Learning to search for structured prediction}
\label{sec:l2s}
Structured prediction is the task of mapping an input to some output with complex internal structure. For example, mapping an English sentence to a sequence of part of speech tags (part of speech tagging), to a syntactic structure (parsing) or to a meaning-equivalent sentence in Chinese (translation). 
Learning to search is a family of approaches for solving structured prediction tasks and encapsulates a number of specific algorithms (e.g., \cite{collins04incremental,daume05laso,xu07beam,xu07planning,ratliff07boosting,daume09searn,ross11dagger,syed11reduction,huang12structured,doppa12oss,doppa14hcsearch}). Learning to search approaches (1) decompose the production of the structure output in terms of an explicit search space (states, actions, etc.); and (2) learn hypotheses that control a policy that takes actions in this search space.

We implemented a learning to search algorithm (based
on~\cite{daume09searn,ross11dagger}) that operates via reduction to
cost sensitive classification which is then further reduced to
regression.  This algorithm was then extensively tested against a
suite of many structured learning algorithms which we report here (see
\cite{daume14imperativesearn} for full details).
The first task we considered was sequence labeling problem: Part of Speech tagging based on data form the Wall Street Journal portion of the Penn Treebank ($45$ labels, evaluated by Hamming loss, $912k$ words of training data). The second is a sequence \emph{chunking} problem: named entity recognition using the CoNLL 2003 dataset ($9$ labels, macro-averaged F-measure, $205k$ words of training data).

We use the following freely available systems/algorithms as points of comparison:
\begin{enumerate}
\item \system{CRF++} The popular \system{CRF++}\ toolkit \cite{crf++} for conditional random fields \cite{lafferty01crf}, which implements both L-BFGS optimization for CRFs \cite{malouf02opt} as well as ``structured MIRA'' \cite{crammer03mira,mcdonald04margin}.
\item \system{CRF SGD} A stochastic gradient descent conditional random field package \cite{crfsgd}.
\item \system{Structured Perceptron} An implementation of the structured perceptron~\cite{collins02perceptron} due to \cite{chang13svm}.
\item \system{Structured SVM} The cutting-plane implementation \cite{joachims09cuttingplane} of the structured SVMs \cite{tsochantaridis04svmiso} for ``HMM'' problems.
\item \system{Structured SVM (DEMI-DCD)} A multicore algorithm for optimizing structured SVMs called DEcoupled Model-update and Inference with Dual Coordinate Descent.
\item \system{VW Search} Our approach is implemented in the Vowpal Wabbit toolkit on top of a cost-sensitive classifier \cite{beygelzimer05reductions} that reduces to regression trained with an online rule incorporating AdaGrad \cite{duchi2011adaptive}, per-feature normalized updates, and importance invariant updates.  The variant \system{VW Search (own fts)} uses computationally inexpensive feature construction facilities available in Vowpal Wabbit (e.g., token prefixes and suffixes), whereas for comparison purposes \system{VW Search} uses the same features as the other systems.
\item \system{VW Classification} An \emph{unstructured} baseline that predicts each label independently, using one-against-all multiclass classification \cite{beygelzimer05reductions}.
\end{enumerate}
These approaches vary both objective function (CRF, MIRA, structured SVM, learning to search) and optimization approach (L-BFGS, cutting plane, stochastic gradient descent, AdaGrad). All implementations are in C/C++, except for the structured perceptron and DEMI-DCD (Java).


\begin{figure}
\centering
\includegraphics[width=3.6in]{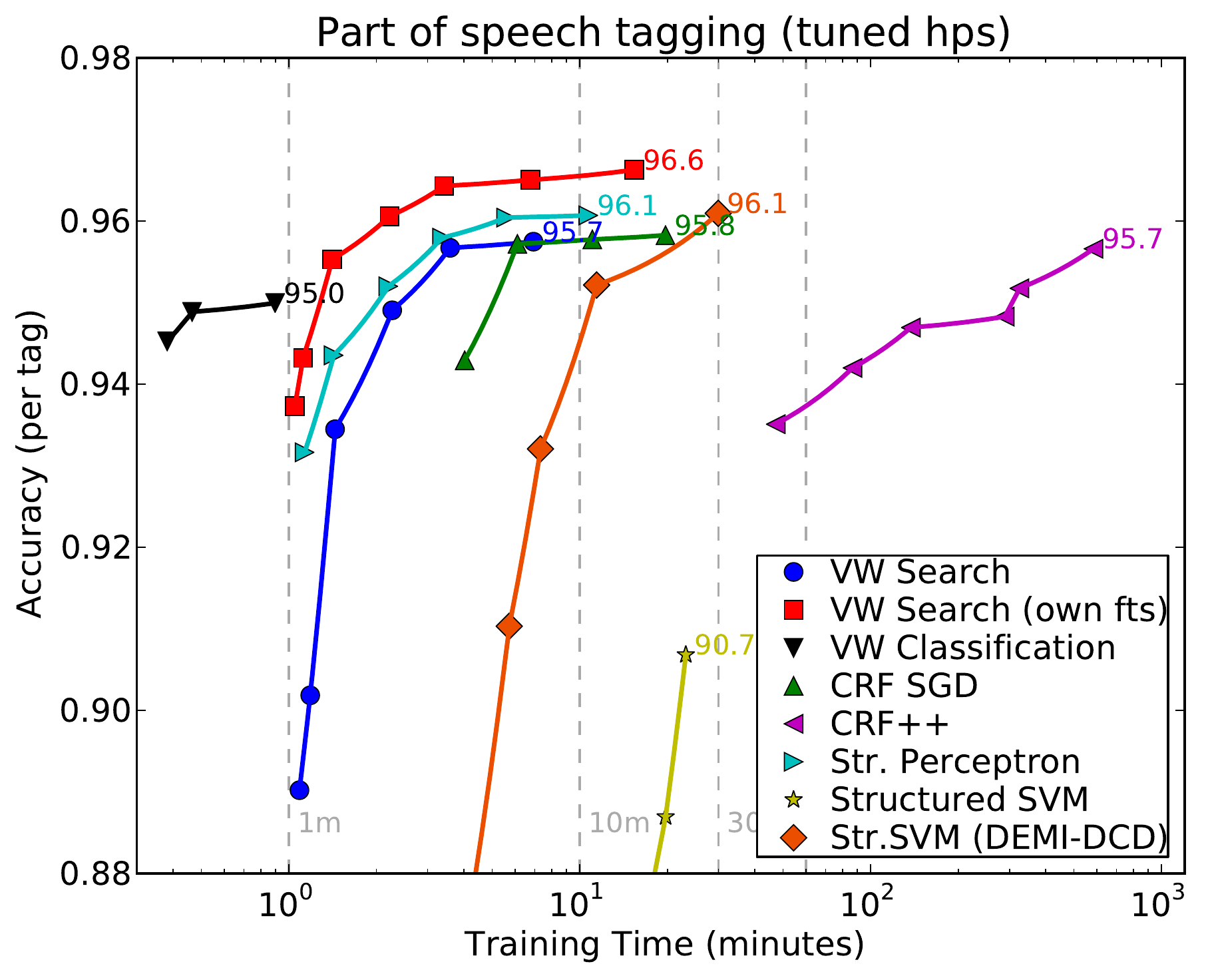} \\

\bigskip
\includegraphics[width=3.6in]{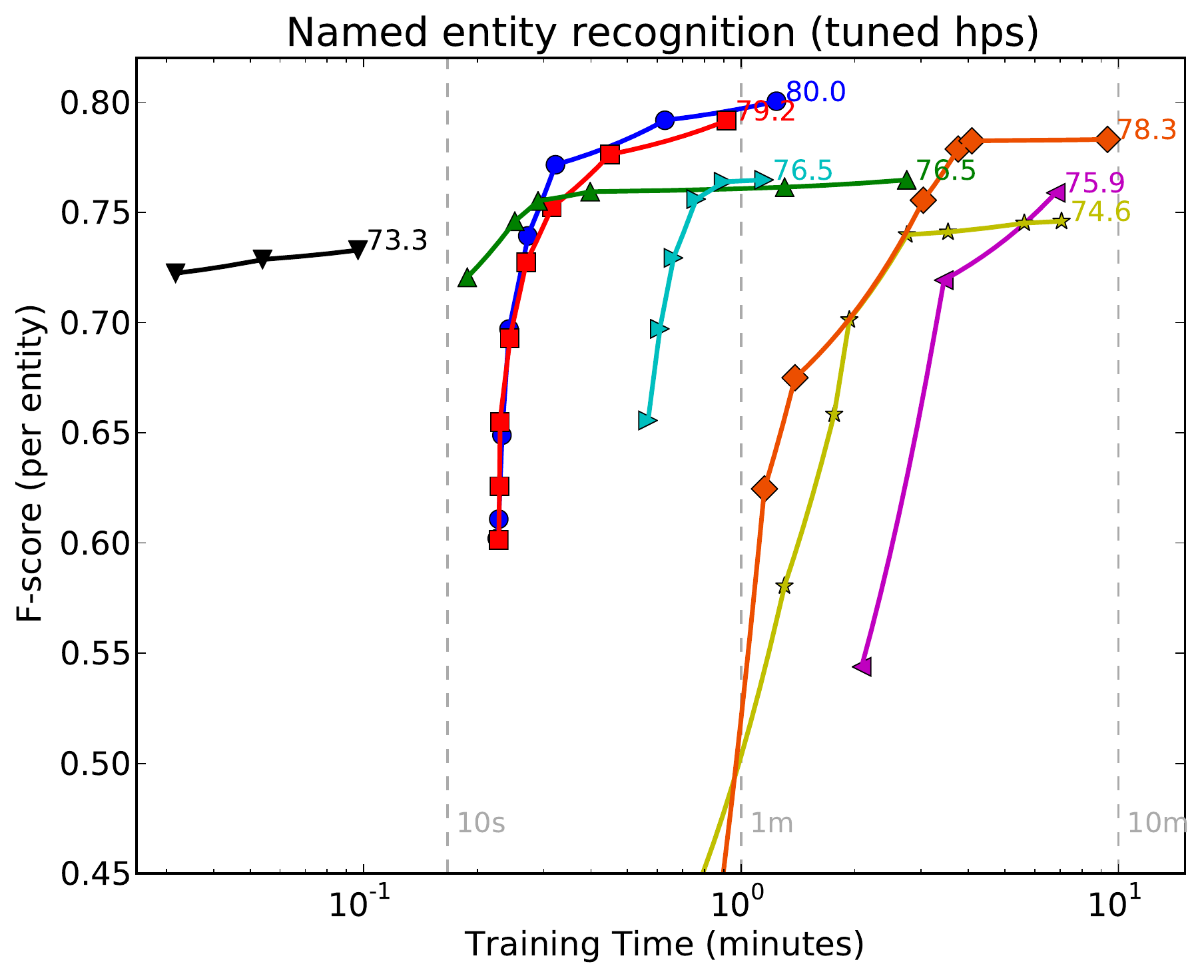}
\caption{Training time versus evaluation accuracy for part of speech tagging (left) and named entity recognition (right). X-axis is in log scale. Different points correspond to different termination criteria for training. Both figures use hyperparameters that were tuned (for accuracy) on the heldout data. (Note: lines are curved due to log scale x-axis.)}
\label{fig:results-tuned}
\end{figure}

In Figure~\ref{fig:results-tuned}, we show trade-offs between training time (x-axis, log scaled) and prediction accuracy (y-axis) for the six systems described previously. The left figure is for part of speech tagging and the right figure is for named entity recognition.
For POS tagging, the independent classifier is by far the fastest (trains in less than one minute) but its performance peaks at $95\%$ accuracy. Three other approaches are in roughly the same time/accuracy tradeoff: \system{VW Search}, \system{VW Search (own fts)} and \system{Structured Perceptron}. All three can achieve very good prediction accuracies in just a few minutes of training. \system{CRF SGD} takes about twice as long. \system{DEMI-DCD} eventually achieves the same accuracy, but it takes a half hour. \system{CRF++} is not competitive (taking over five hours to even do as well as \system{VW Classification}). \system{Structured SVM} (cutting plane implementation) runs out of memory before achieving competitive performance, likely due to too many constraints.

For NER the story is a bit different. The independent classifiers are far from competitive. Here, the two variants of \system{VW Search} totally dominate. In this case, \system{Structured Perceptron}, which did quite well on POS tagging, is no longer competitive and is essentially dominated by \system{CRF SGD}. The only system coming close to \system{VW Search}'s performance is \system{DEMI-DCD}, although its performance flattens out after a few minutes.\footnote{We also tried giving \system{\scriptsize CRF SGD} the features computed by \system{VW Search (own fts)} on both POS and NER. On POS, its accuracy improved to 96.5---on par with \system{VW Search (own fts)}---with essentially the same speed. On NER its performance decreased. For both tasks, clearly features matter. But which features matter \emph{is} a function of the approach being taken.}

In addition to training time, test time behavior can be of high
importance in natural applications.  On NER, prediction times varied from
$5.3k$ tokens/second (\system{DEMI-DCD} and \system{Structured
  Perceptron} to around $20k$ (\system{CRF SGD} and \system{Structured
  SVM}) to $100k$ (\system{CRF++}) to $220k$ (\system{VW (own fts)})
and $285k$ (\system{VW}). Although \system{CRF SGD} and
\system{Structured Perceptron} fared well in terms of training time,
their test-time behavior is suboptimal.

When looking at POS tagging, the effect of the $\mathcal{O}(k)$ dependence on the size of the label set further increased the (relative) advantage of \system{VW Search} over alternatives.

\section{Summary and future directions}

In working with learning reductions for several years, the greatest
benefits seem to be incurred with modularity, deeper reductions, and
computational efficiency.  

Modularity means that the extra code required for multiclass
classification (for example) is minor compared to the code required
for binary classification.  It also simplifies the use of a learning
system, because (for example) learning rate flags apply to all
learning algorithms.  Modularity is also an easy experimentation
and optimization tool, as one can plug in different black 
boxes for different modules.

While there are many experiments showing near-parity prediction
performance for simple reductions, it appears that for deeper
reductions the advantage may become more
pronounced.  This is well illustrated for the learning to search
results discussed in section~\ref{sec:l2s}, but has been observed with
contextual bandit learning as well~\cite{minimonster}.  The precise
reason for this is unclear, as it is very difficult to isolate the
most important difference between very different approaches to solving
the problem.

Not all machine learning reductions provide computational benefits,
but those that do may provide enormous benefits.  These are
mostly detailed in section~\ref{sec:unique}, with benefits often
including an exponential reduction in computational complexity.  

In terms of the theory itself, we have often found that qualitative
transitions from an error reduction to a regret reduction are
beneficial.  We have also found the isolation of concerns via
encapsulation of the optimization problem to be quite helpful in
developing solutions.

We have not found that precise coefficients are predictive of relative
performance amongst two reductions accomplishing the same task with
the same base learning algorithm but different representations.  As an
example, the theory for error correcting tournaments~\cite{ECT} is
substantially stronger than for one-against-all, yet often
one-against-all performs better empirically.  The theory is of course
not wrong, but since the theory is relativized by the performance of
the base predictor, the representational compatibility issue can and
does play a stronger role in predicting performance.

There are many questions we still have about learning reductions.
\begin{enumerate}
\item Can the interface we have support effective use of SIMD/BLAS/GPU
  approaches to optimization?  Marrying the computational benefits of
  learning reductions to the computational benefits of these
  approaches could be compelling.
\item Is the learning reduction approach effective when the base
  learner is a multitask (possibly ``deep'') learning system?  Often
  the different subproblems created by the reduction share enough
  structure that a multitask approach appears plausibly effective.
\item Can the learning reduction approach be usefully applied at the
  representational level?  Is there a theory of representational
  reductions?
\end{enumerate}

\end{document}